\documentclass[journal,twoside,web]{ieeecolor}
\usepackage{jsen}
\usepackage{cite}
\usepackage{amsmath,amssymb,amsfonts}
\usepackage{algorithmic}
\usepackage{graphicx}
\usepackage{textcomp}
\usepackage{wrapfig}
\def\BibTeX{{\rm B\kern-.05em{\sc i\kern-.025em b}\kern-.08em
    T\kern-.1667em\lower.7ex\hbox{E}\kern-.125emX}}
\definecolor{abstractbg}{rgb}{0.89804,0.94510,0.83137}
\begin{document}
\title{SCTracker: Multi-object tracking with shape and confidence constraints}
\author{Huan Mao, Yulin Chen, Zongtan Li, Feng Chen, \IEEEmembership{Member, IEEE}, and Pingping Chen, \IEEEmembership{Member, IEEE}
\thanks{This work was supported by the National Natural Science Foundation of China (Grant Nos. 61871132).}
\thanks{Huan Mao, Yulin Chen, Zongtan Li, Feng Chen, and Pingping Chen are with the College of Physics and Information Engineering, Fuzhou University, Fuzhou 350108, China(e-mail: maohuan980202@gmail.com; 211127057@fzu.edu.cn; zongtanli2023@163.com; chenf@fzu.edu.cn; ppchen.xm@gmail.com).}
}

\IEEEtitleabstractindextext{%
\fcolorbox{abstractbg}{abstractbg}{%
\begin{minipage}{\textwidth}%
\begin{wrapfigure}[12]{r}{3in}%
\includegraphics[height=1.4in,width=2.8in]{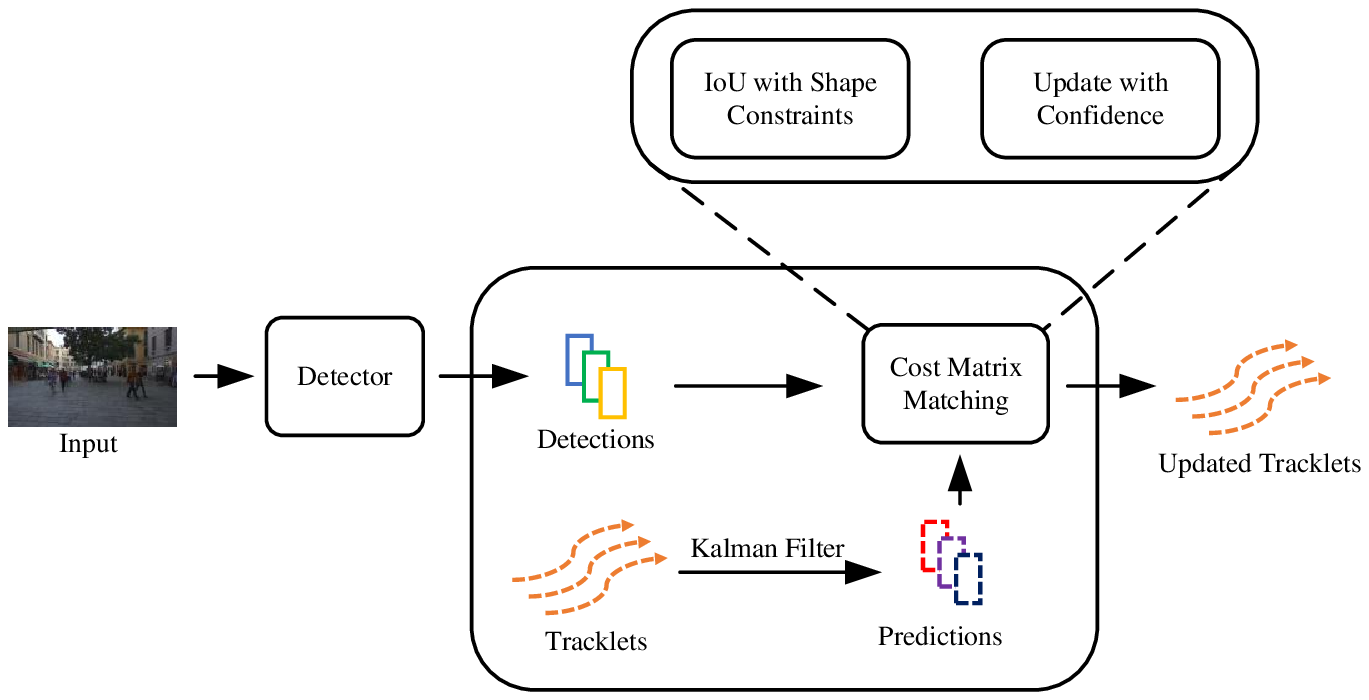}%
\end{wrapfigure}%
\begin{abstract}
Detection-based tracking is one of the main methods of multi-object tracking. It can obtain good tracking results when using excellent detectors but it may associate wrong targets when facing overlapping and low-confidence detections.
To address this issue, this paper proposes a multi-object tracker based on shape constraint and confidence named SCTracker. In the data association stage, an Intersection of Union distance with shape constraints is applied to calculate the cost matrix between tracks and detections, which can effectively avoid the track tracking to the wrong target with the similar position but inconsistent shape, so as to improve the accuracy of data association.
Additionally, the Kalman Filter based on the detection confidence is used to update the motion state to improve the tracking performance when the detection has low confidence.
Experimental results on MOT 17 dataset show that the proposed method can effectively improve the tracking performance of multi-object tracking.

\end{abstract}

\end{minipage}}}

\maketitle

\section{Introduction}
\label{sec:introduction}
\IEEEPARstart{M}{ulti}-object tracking is widely used in video surveillance, behavior analysis, traffic monitoring and other fields\cite{luo2021multiple}. Therefore, it has important theoretical research significance and practical application value.
According to the different sensors used, it can be divided into visual multi-object tracking, radar multi-object tracking, multi-object tracking with multi-sensor fusion, and so on. Among them, visual multi-object tracking is the mainstream direction in the field of multi-object tracking.
Visual multi-object tracking aims to detect multiple objects in the video sequence and assign identification numbers to them.
With the rapid development of deep learning and multi-object detection, multi-object tracking based on detection has become a research hotspot in recent years\cite{bochinski2017high, bewley2016simple, wojke2017simple, zhang2022bytetrack}. Multi-object tracking based on detection is mainly composed of a detector and data association. First, a well trained detector is used to obtain detection results. In the data association stage, detections are associated with tracks according to some specific measures and association strategies. When the same detector is used, the data association algorithm determines whether the multi-object tracking algorithm can assign the correct ID number to the target in different scenarios.

Most researches on multi-object tracking are divided into detection-based tracking and end-to-end tracking. The performance of Detection-based tracking depends largely on the performance of the detector. In the method of using Re-identification (ReID) features, the ReID feature extraction network brings additional training and inference costs. Some methods\cite{wang2020towards, zhang2021fairmot} try to integrate ReID feature extraction into the detector for joint training, but there is a problem of imbalance between the detection branch and the feature extraction branch. Recently, end-to-end multi-object tracking methods\cite{chu2023transmot, sun2020transtrack, xu2021transcenter, meinhardt2022trackformer, zeng2022motr} based on Transformer\cite{vaswani2017attention} have attracted more attention. End-to-end tracking needs to realize detection and tracking in a network, but its performance still lags behind the tracking based on detection.

Current target association methods mainly measure the distance between the track and the detection box from the spatial information and re-identification features. These method use the Hungarian algorithm to assign the most similar detection box for the track, and finally completes the update of the track.
However, due to the lack of differentiation in the Intersection of Union (IoU) distance measurement, the position information becomes unreliable when dealing with occlusions and low confidence detection boxes.
Although the re-identification features can effectively find the long-disappeared object, the spatial motion information is ignored, and the feature extraction network also brings additional training and time costs. In addition, how to balance the feature and location information is also a problem brought by the introduction of re-identification features.
To solve the above problems and challenges, this paper proposes a multi-object tracker based on shape constraint and confidence named SCTracker. SCTracker uses an Intersection of Union distance with shape constraints (Shape-IoU), including a height constraint and an area constraint, to calculate the cost matrix. The measurement of different shapes but similar IoU in the track and detections is optimized to improve the accuracy of data association. The Kalman filter parameters of the tracklet are updated with the confidence of the detection in order to improve the tracking performance when the detection result is poor or the bounding box mutates.

The main contributions of this paper are summarized as follows:
\begin{enumerate}
\item An Intersection of Union distance with shape constraints (Shape-IoU) is proposed to distinguish the detection results of different shapes but the same intersection distance with tracks.
\item A new track state update based on the confidence of detections is proposed in order to better describe the motion state of the tracklet associated with the low quality detection result.
\item Compared with other advanced methods in MOT 17 dataset\cite{milan2016mot16}, SCTracker shows better tracking performance and verifies the effectiveness of the proposed method.
\end{enumerate}

\section{Related Work}\label{sec2}
With the development of deep learning in recent years, the improvement of multi-object detection performance enables detection-based tracking methods to obtain strong tracking performance.
The detection-based tracking method describes the state of the track through feature extraction, motion prediction and other methods.
IoU Tracker\cite{bochinski2017high} directly associates objects by the IoU distance of detection results between adjacent frames to achieve high running speed, but the tracking performance is prone to be affected by occlusion scenes.
SORT\cite{bewley2016simple} uses the Kalman Filter for motion prediction, effectively overcoming the problem of track loss in short-time occlusion scenes.
On this basis, Deep SORT\cite{wojke2017simple} adds an appearance feature and cascade matching strategy to improve tracking loss under long-term occlusion.
ByteTrack\cite{zhang2022bytetrack} reserves detections with low score for a twice association strategy, and only uses the Kalman Filter to predict the motion state of the track to achieve the most advanced tracking performance with a stronger detector.
QDTrack\cite{pang2021quasi} intensively samples the surrounding area of the target to obtain rich feature information. In the inference stage, only the feature information is used to carry out bidirectional matching between two frames to achieve tracking.

Zhongdao et al.\cite{wang2020towards} extended an appearance feature branch on the single-stage detector YOLO v3\cite{redmon2018yolov3}, jointly learning detection and appearance embedding in the way of multi-task learning, reducing the complexity of the algorithm. However, there is a conflict between the detector and feature learning that learning high-dimensional feature embedding caused imbalance between detection branch and feature branch.
FairMOT\cite{zhang2021fairmot} uses CenterNet\cite{zhou2019objects}, a detector based on the target center point, to extend the low-dimensional feature branch to balance the multi-task learning of detection and feature extraction, which improves the tracking performance while maintaining high reasoning speed.

End-to-end tracking methods based on Transformer\cite{vaswani2017attention} have been concerned in some recent studies\cite{chu2023transmot, sun2020transtrack, xu2021transcenter, meinhardt2022trackformer, zeng2022motr}. They are based on the normal form of DETR\cite{carion2020end, zhu2020deformable} to build the query vector of tracks, which can implicitly build the object trajectories and achieve end-to-end multi-object tracking. However, the performance of end-to-end tracking still lags behind that of detection-based tracking, and the calculation speed is slow.

All the above multi-object tracking methods have achieved excellent tracking performance. However, the tracking performance under complex scenes and low quality detection results still needs to be improved. To solve these problems, we propose SCTracker based on detection-based tracking paradigm to improve the simple measurement of distance and the track update under low-confidence detection results, so as to improve the tracking performance while maintaining high speed.

\section{Proposed method}\label{sec3}

\subsection{Overall framework}\label{subsec3_1}

\begin{figure*}[h!t]
\centering
\includegraphics[width=0.9\textwidth]{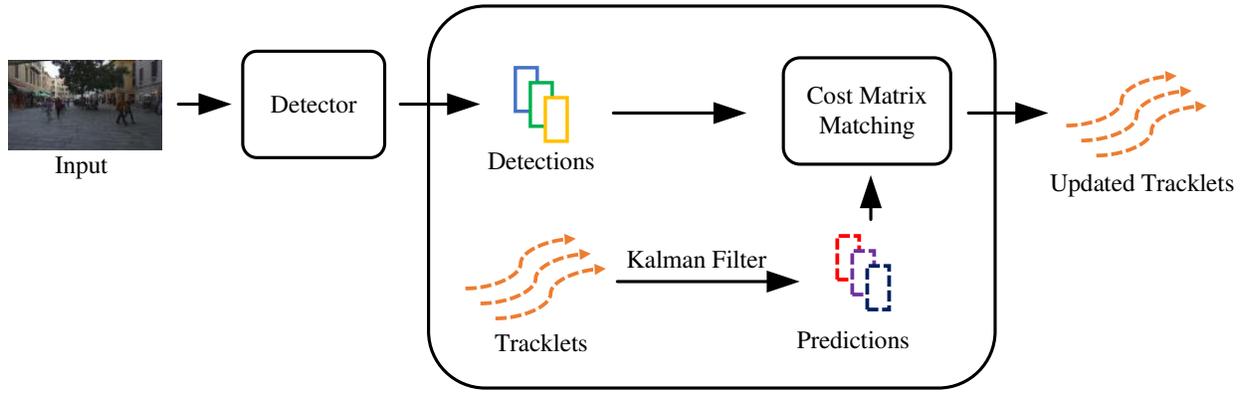}
\caption{The pipeline of the proposed method}\label{fig3_1}
\end{figure*}

 The proposed SCTracker only uses the spatial information of detections in the data association stage without using the re-identification feature in order to avoid the extra training cost and improve the efficiency of the algorithm.
The Kalman Filter is applied to predict the motion state of a track, which is described as an 8-dimensional vector $[x, y, a, h, d_x, d_y, d_a, d_h]$, where $(x, y, a, h)$ are the coordinates of the top left corner of the track, the aspect ratio and the height of the frame respectively, and the posterior four dimensions represent the speed of the first four dimensions of the state vector.
 The overall process of the proposed tracker based on shape constraint and confidence is shown in Fig. \ref{fig3_1}.

Specifically, the image of frame $i$ is input into the detector to obtain a set of detections $Detections(t)=\{B^t_1, ..., B^t_N\}$ with a total number of $N$, in which each detection contains four components $(x, y, a, h)$, which are the top-left coordinate of the detection, the aspect ratio and the height of the box, respectively.
The set of tracks in the current frame participating in the data association is represented as $Tracks(t)=\{S^t_1, ..., S^t_M\}$, where $M$ is the number of tracks currently participating in the association.

%\subsubsection{Data association}\label{subsubsec3_1_1}

The association strategy is built on ByteTrack\cite{zhang2022bytetrack}, which considers the information of detections with low confidence, and detections are divided into a set of high confidence detections $Detections_{high}(t)\in Detections(t)$ and a set of low confidence detections $Detections_{low}(t)\in Detections(t)$.
Specifically, for each 8-dimensional state vector, the Kalman Filter is applied to predict the state distribution in the current frame, and the bounding box of the tracklet in the current frame is set as the first four-dimensional component $(x',y',a',h')$ of the predicted state vector.

In the first association, the distance between the set of tracks $Tracks(t)$ and the set of detections with high confidence $Detections_{high}(t)$ is calculated to obtain a cost matrix $C \in \mathbb{R}^{M \times N}$. According to this cost matrix, the Hungarian algorithm is used to assign the corresponding detection to the track. Unmatched detections are created as new tracks, and unmatched tracks are represented as $Tracks_{remain}(t)$ and participate in the second association.

The distance between $Tracks_{remain}(t)$ and detections with low confidence is calculated in the second association and the assignment is calculated by the Hungarian algorithm. Track marks that are not matched in the second association are lost. The track that is not matched in the second association is marked as a lost state, and the track whose lost state exceeds a certain number of frames is marked as a removed state and does not participate in the tracking of subsequent frames. Finally, the Kalman filter parameters of matched tracks are updated and the tracking result of the current frame is returned.

The calculation of the cost matrix of the above two correlations needs to measure the spatial distance between tracks and detections, so the cost matrix calculated by M tracks and N detections are formulated as:

\begin{equation}
C=\begin{bmatrix}
d_{11}  & \dots     & d_{1N}\\
\vdots  & \ddots    & \vdots\\
d_{M1}  & \dots     & d_{MN}
\end{bmatrix}_{M \times N},
\end{equation}
where $d_{ab} = d(B_a, B_b)$ is the spatial distance between bounding box $B_a$ and bounding box $B_b$. In this paper, Shape-IoU is used to calculate the distance $d(B_a, B_b)$. For the matched tracks, the Kalman filter parameters are updated based on the confidence of the corresponding detection.

 \subsection{IoU with shape constraints}\label{subsec3_2}

 In the process of data association, it is necessary to measure the distance between the detection and the track to get the cost matrix, and then use the Hungarian algorithm to assign the track of the detection with the minimum cost. In most detection-based tracking methods, the IoU distance is used to measure the overlap of detection and track. Specifically, IoU is the ratio of the intersection area to the union area of two bounding boxes, which measures the overlapping degree between the two boxes. The function expression of the IoU is as follows:

\begin{equation}
IoU(B_a, B_b) = \frac{| B_a \cap B_b |}{| B_a \cup B_b |}.
\end{equation}

The IoU between the two bounding boxes is 1 when they are completely overlapped, and 0 when they are not. Therefore, the IoU distance is expressed as follows:

\begin{equation}
d_{IoU}(B_a, B_b) = 1 - IoU(B_a, B_b).
\end{equation}

\begin{figure}[h]
\centering
\includegraphics[width=0.4\textwidth]{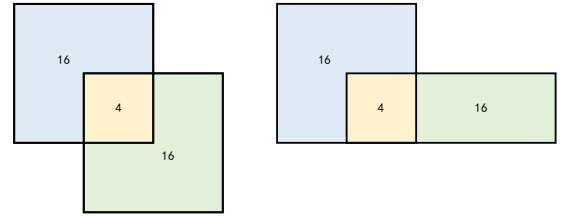}
\caption{Schematic diagram of different shape bounding boxes intersecting the same bounding box}\label{fig3_2}
\end{figure}

In the multi-object tracking, predicted bounding boxes of tracks and ones of detections under the current frame are involved in the calculation of IoU distance. However, detections with different shapes and sizes may have the same IoU distance with the same predicted box of the track. As shown in Fig. \ref{fig3_2}, green bounding boxes of different shapes have the same intersection area and union area with the same blue bounding box, but the shapes of the two green bounding boxes are completely different.
Tracking targets are mainly objects with relatively fixed shapes, such as pedestrians or vehicles, and detections with inconsistent shapes are mostly false positive detection results in multi-object tracking. Therefore, it is difficult to accurately measure the distance between targets and tracks using only IoU distance, which may cause the wrong matching results of the tracker.

In order to solve the above problems in multi-object tracking, we propose an Intersection of Union distance with shape constraints named Shape-IoU. By adding constraints to the shape of the detection, the distance between the track and the false positive detection frame with inconsistent shape but large IoU becomes larger, and the distance between the track and the detection with consistent shape and large intersection ratio is reduced.
The shape constraint item contains a height constraint item $\rho_h$ and an area constraint item $\rho_s$. Intuitively, the height of the target is less affected by complex scenes, occlusion and other factors, so adding a height constraint item can determine the shape of a detection result according to the area and height. The Shape-IoU is formulated as:

\begin{equation}
d(B_a, B_b) = 1 - IoU(B_a, B_b) + \rho_h(B_a, B_b) + \rho_s(B_a, B_b),
\end{equation}
\begin{align}
\rho_h(B_a, B_b) &= \frac{(h_a - h_b)^2}{(h_u + \epsilon)^2} ,  \\
\rho_s(B_a, B_b) &= \frac{(S_a - S_b)^2}{(S_u + \epsilon)^2} ,
\end{align}
where $h_a, h_b, h_u$ is the height of the two boundary boxes and the corresponding minimum enclosing rectangle respectively, $S_a, S_b, S_u$ is the area of the two boundary boxes and the corresponding minimum enclosing rectangle respectively, $\epsilon$ is a valid minimum to ensure the validity of the constraint term, and $\epsilon$ is set as $10^{-7}$ in the experiment.

\subsection{Track update based on detection confidence}\label{subsec3_3}

\begin{figure*}[h]
\centering
\includegraphics[width=0.9\textwidth]{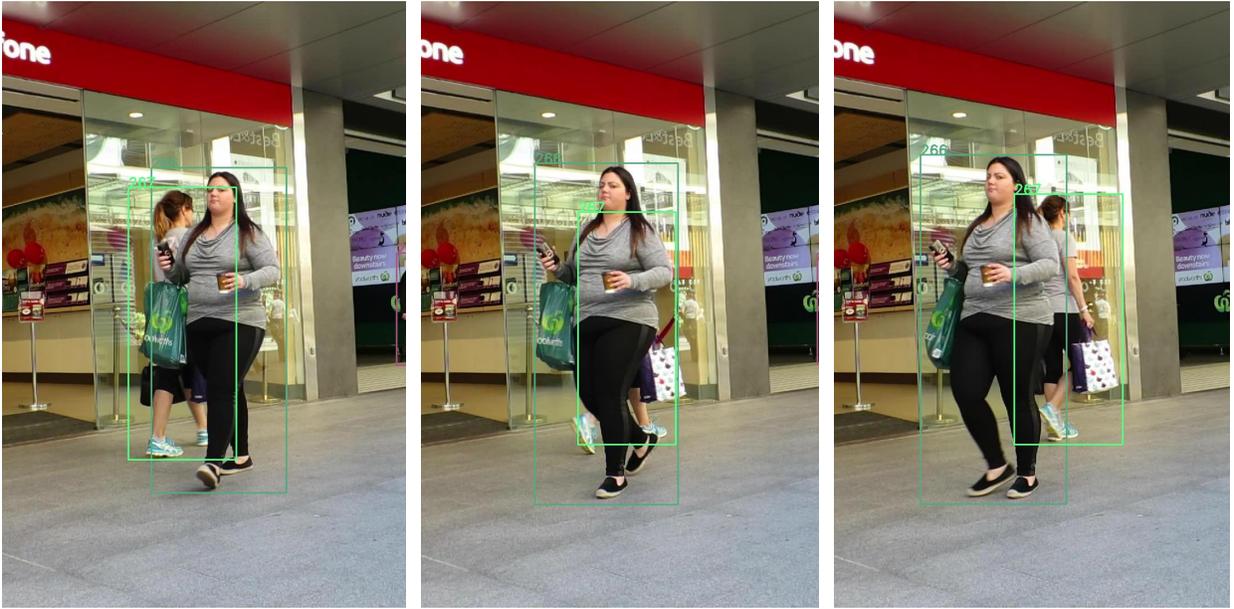}
\caption{Visualization results of low confidence detection box}\label{fig3_3}
\end{figure*}

In the tracking process, the Kalman Filter is used to build a motion model and predict the distribution of the next position for the tracklet, which can effectively avoid the tracking loss caused by short-time occlusion and enhance the tracking performance.
Although ByteTrack\cite{zhang2022bytetrack} takes into account the effective information of low score detection box, the detection quality of detector output deteriorates due to occlusions, environmental interference and other factors, and the detection results with low confidence have a certain deviation from the ground truth, which results in the prediction error of the updated state vector for the next frame and the error accumulation in subsequent frames. As shown in Fig. \ref{fig3_3}, when the person behind is blocked, the size and shape of its bounding box are inaccurate, and there is a certain error with the actual size and shape.

In view of the above problems, we propose a track update strategy based on detection confidence. Specifically, the measurement noise covariance matrix $R$ in the update equations describes the uncertainty in the measurement process, which is related to the confidence of the detection results output by the detector in the tracking task. Therefore, the confidence of detection is added as a weighting factor, and the specific function expression is as follows:

\begin{equation}
R_c = R \cdot (1 - score^2),
\end{equation}
where $score$ is the confidence of the detection result and the measurement noise covariance matrix $R$ is negatively correlated with $score$.
In addition, in the case of short-time occlusion, the shape of the detection output by the detector appears to be biased. To avoid the influence of incorrect shape changes on the state vector of the Kalman filter, the confidence of detection is introduced to correct the velocity component of the state vector when the original state vector $m$ is updated to a new state vector $m'$ by using the Kalman filter, as shown in the following formula:

\begin{equation}
m_{vel}' = score \cdot m_{vel}' + (1 - score) \cdot m_{vel},
\end{equation}
where, $m_{vel}, m_{vel}'$ is respectively the last four dimensional vector of the original state vector $m$ and the updated state vector $m'$. When the confidence is high, the updated velocity component is more inclined. On the contrary, when the confidence is low, it is considered that the updated velocity component has a larger error and is more inclined to the original state vector velocity component.

\section{Experiments and results}\label{sec4}

\subsection{Datasets description}\label{subsec4_1}

 \begin{figure}[h]
\centering
\includegraphics[width=0.4\textwidth]{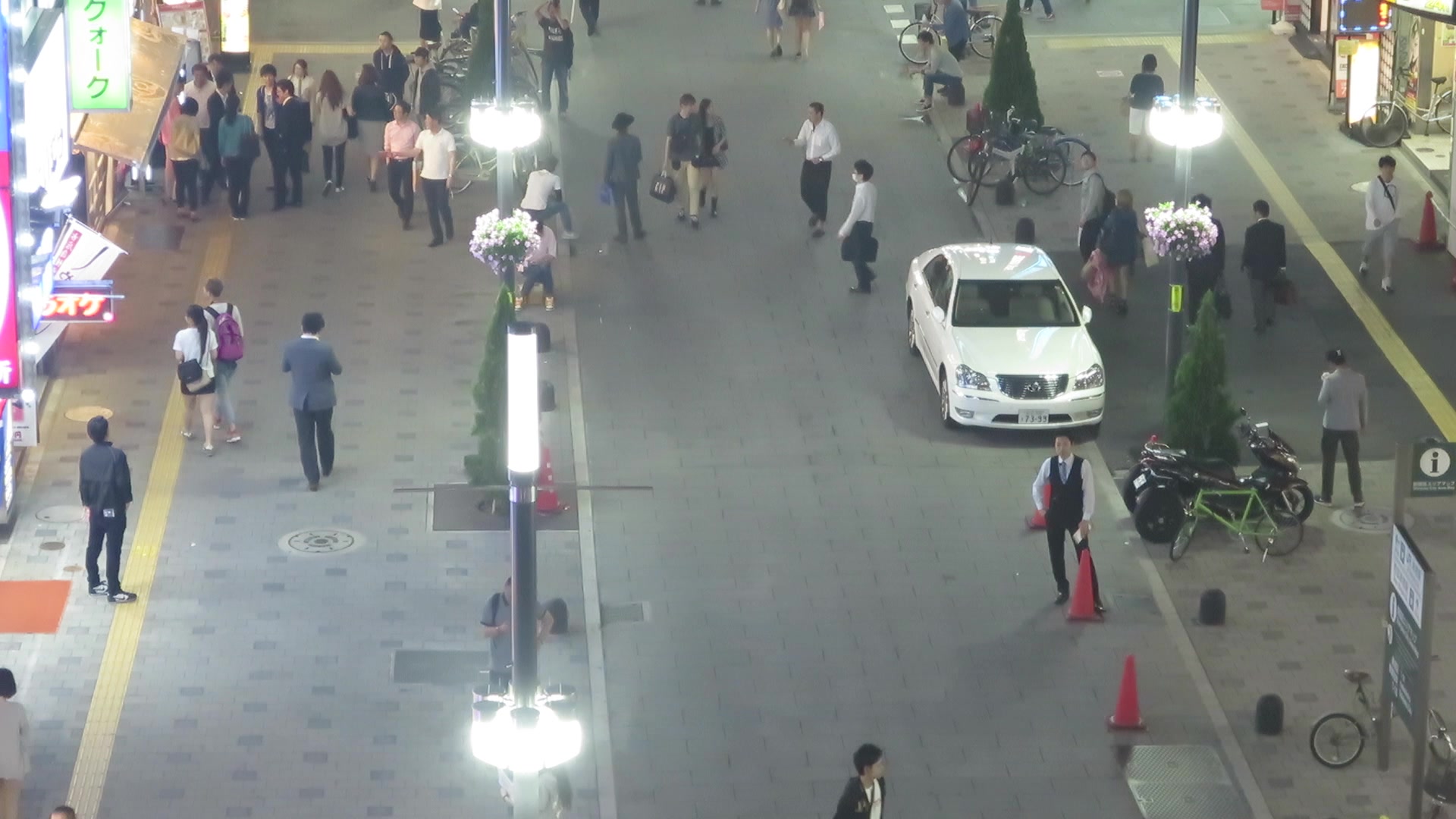}
\caption{An example of image in MOT 17 dataset}\label{fig4_1}
\end{figure}

In order to evaluate the performance of the proposed method, MOT 17\cite{milan2016mot16} is used as a unified benchmark dataset for experiments. MOT 17 is a widely used multi-object tracking dataset with large scale, authenticity and high quality. The dataset comprises of 17 video sequences of real-world scenes shot by different cameras, including different scenes such as shopping malls, streets and subway stations. Each video sequence has a resolution of 720p, a frame rate of 30 frames per second, and contains a different number and variety of targets.
In order to facilitate the evaluation of the experiment, the first half of each video sequence in the training set of MOT 17 is divided as training, and the second half is used as the validation set of MOT 17 to test the performance of the model.
Fig. \ref{fig4_1} shows an example image of the MOT 17 dataset.

\subsection{Evaluation metrics}\label{subsec4_2}

In order to make the performance of this algorithm comparable with other algorithms, all experiments follow the CLEAR Multi-object metrics\cite{bernardin2008evaluating}, including MOTA, IDF1 and IDSW. MOTA is the multiple object tracking accuracy which reflects the accuracy of determining the number of targets and related attributes of targets. It is used to calculate the error accumulation in tracking, including FP, FN, and IDSW. IDF1 is a comprehensive indicator that considers both the accuracy and stability of the tracker. It is the harmonic mean of MOTA and IDP. IDSW is the Number of Identity Switches.

\subsection{Experiment setting and result}\label{subsec4_3}

The detector in the proposed method uses YOLO X\cite{ge2021yolox}. The model of the detector uses a pre-trained model trained on the COCO dataset\cite{lin2014microsoft} as initialization weight. The input image size is 1440×800. SGD optimizer is used for training 80 epochs. The weight decay and momentum are set to 0.0005 and 0.9 respectively. The initial learning rate is set to 0.001, and the learning rate adjustment strategy uses learning rate preheating and cosine annealing. The inference of the model is performed on an NVIDIA GeForce RTX 3090 GPU, and the inference time consists of the forward propagation time including post-processing and tracking time.

\subsubsection{Ablation experiment}\label{subsubsec4_3_1}

Since there are no learnable parameters in the data association, in the ablation experiment, the detector selects a model of the same size (YOLO X-X) and uses the same training weight. The proposed method is built on ByteTrack\cite{zhang2022bytetrack}, and ByteTrack is chosen as a baseline for comparison. The Intersection of Union distance with shape constraints (Shape-IoU) is denoted as $Shape$, and the track update based on detection confidence is denoted as $Conf$. The experimental results is shown in Table \ref{tab4_1}. Bold indicates the best result in that column.
Under the premise of the same detection results, Shape-IoU compared with the baseline has improved in IDF1 and MOTA indicators, and IDSW has decreased. Compared with the baseline, the track update based on detection confidence has improved by 0.2\% in IDF1, and IDSW has slightly increased. The confidence in using Shape-IoU and the track update based on detection has improved in all metrics. IDF1 has increased by 0.8\%, MOTA has increased by 0.3\%, and IDSW has decreased by 16. For inference time, since the same detector model is used, the forward propagation time of the model remains basically the same. The tracking time of each frame of the baseline association algorithm is about 6 ms, and one of the proposed method is about 12 ms, demonstrating that each component significantly improves multi-object tracking performance at a minimal time cost.

\begin{table}[h]
\center
\caption{Results of ablation experiments on the MOT 17 validation set}\label{tab4_1}
\begin{tabular}{@{}llll@{}}
\hline
Methods     & IDF1\%  & MOTA\% & IDSW\\
\hline
Baseline\cite{zhang2022bytetrack}            & 79.4  &   76.4  &	165  \\
Ours($Shape$)       & 79.9  &	\textbf{76.8}  &	160  \\
Ours($Conf$)        & 79.6  &	76.4  &	168  \\
Ours($Shape+Conf$)  & \textbf{80.2}  &	76.7  &	\textbf{149}  \\
\hline
\end{tabular}
\end{table}

When measuring the distance between tracks and detections, adding different shape constraints affects the cost matrix differently. Therefore, it is necessary to set the ablation of different shape constraints and choose the appropriate term to more accurately describe the distance between tracks and detections. The group experimentally set up height constraint $\rho_h$ and area constraint $\rho_s$, respectively, representing the height and the area distance of two boxes. The experimental results of different shape constraints are shown in Table \ref{tab4_2}, where bold indicates the best result in that column. As shown in Table \ref{tab4_2}, using only height or area constraint can improve IDF1 and MOTA compared with not using shape constraint, and IDSW is also reduced. Using both height and area constraint can maximize the improvement of various tracking metics, fully demonstrating the effectiveness of combining shape constraints to improve multi-object tracking performance.

\begin{table}[h]
\center
\caption{Experimental results of different shape constraint terms}\label{tab4_2}
\begin{tabular}{@{}lllll@{}}
\hline
$\rho_h$ & $\rho_s$ & IDF1\%  & MOTA\% & IDSW\\
\hline
-           &-             &79.4	&76.4	&165  \\
\checkmark  &-             &79.8	&76.5	&\textbf{160}  \\
-           &\checkmark    &79.8	&\textbf{76.8}	&163  \\
\checkmark  &\checkmark    &\textbf{79.9}	&\textbf{76.8}	&\textbf{160}  \\
\hline
\end{tabular}
\end{table}

\subsubsection{Comparative experiment}\label{subsubsec4_3_2}

 \begin{figure*}[h!t]
\centering
\includegraphics[width=0.9\textwidth]{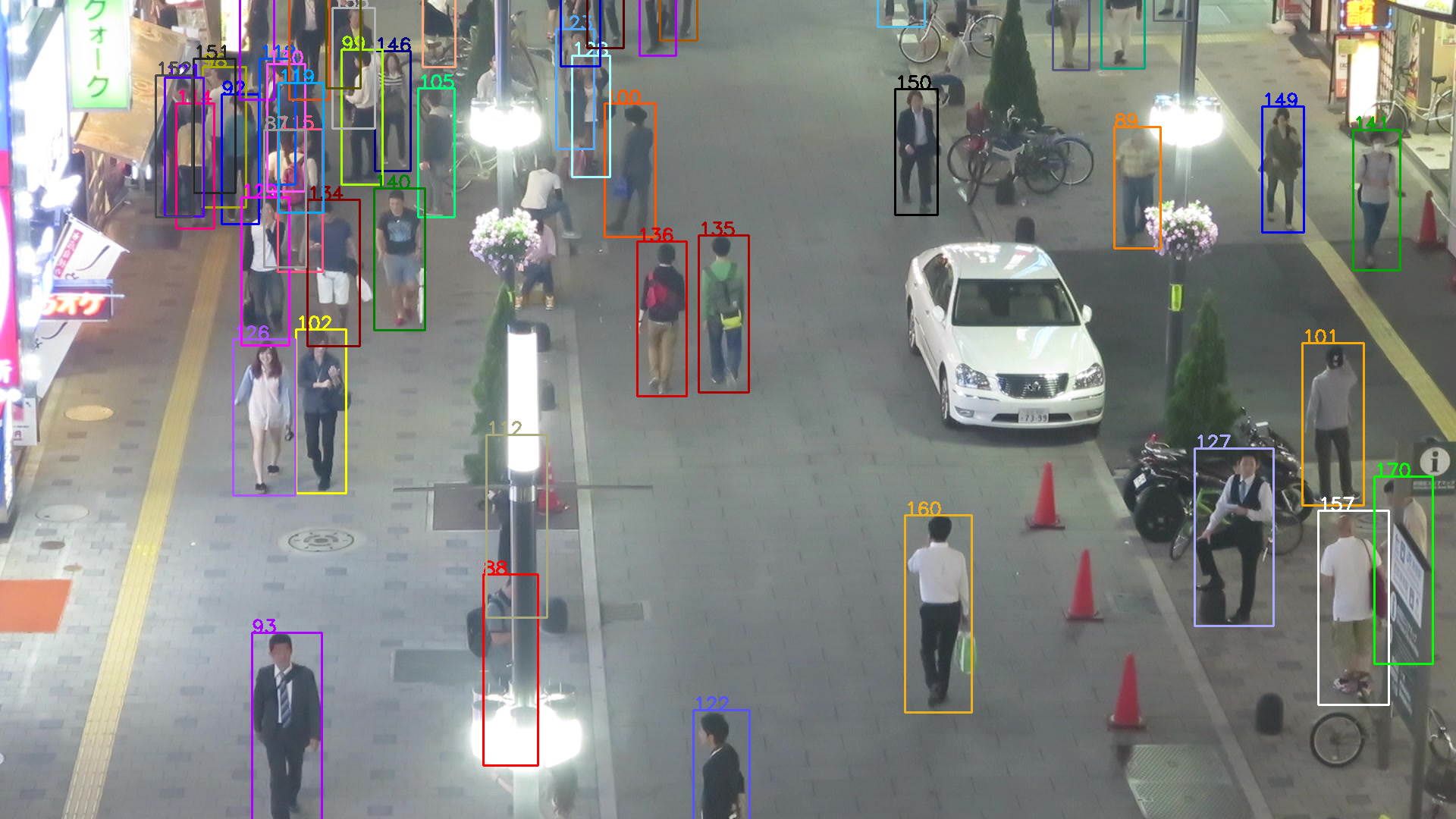}
\caption{Visualization results of the proposed method on the MOT 17}\label{fig4_2}
\end{figure*}

\begin{table}[h]
\center
\caption{Results of different methods on the MOT 17 validation set}\label{tab4_3}
\begin{tabular}{@{}llll@{}}
\hline
Methods     & IDF1\%  & MOTA\% & IDSW\\
\hline
JDE\cite{wang2020towards}       & 65.0  &	60.3  &	474  \\
FairMOT\cite{zhang2021fairmot}        & 71.8  &	67.7  &	463  \\
Baseline\cite{zhang2022bytetrack}            & 79.4  &   76.4  &	165  \\
Ours  & \textbf{80.2}  &	76.7  &	\textbf{149}  \\
\hline
\end{tabular}
\end{table}

The proposed method is compared with other excellent multi-object tracking methods. The performance of each method on the MOT 17 validation set is evaluated, and the advantages of the proposed algorithm are illustrated through intuitive experimental results. The results of each performance metric are presented in tabular form, where each bold part in each column represents the best metric value. Table \ref{tab4_3} shows the experimental results of different multi-object tracking methods on the MOT 17 validation set. It can be observed that the IDF1 of the proposed method is 80.2\%, MOTA is 76.7\%, and IDSW is 149, all of which are better than the experimental results of other multi-object tracking methods. The visualization results of the algorithm proposed in this chapter on MOT 17 are shown in Fig. \ref{fig4_2}.

\section{Conclusion}\label{sec5}

In this paper, a multi-object tracker based on shape constraint and confidence named SCTracker is proposed to optimize the data association of detections with different shapes and improve the tracking with low quality detection results. First, in view of the shape similarity between detections and tracks, constraints of height and area are added to the calculation of IoU distance to avoid the influence of false association of detection frames with similar positions but inconsistent shapes. Second, the confidence of the detections was introduced in the Kalman Filter update of tracklets. The measurement noise covariance matrix based on the confidence of detections is used to update and the velocity component of the state vector is modified according to the confidence of detections, improving the tracking performance of low quality detection results. Finally, the performance of the proposed method and the ablation of each component were evaluated on the verification set of MOT 17. The experimental results fully show that the algorithm has advanced tracking performance.

\begin{IEEEbiography}[{\includegraphics[width=1in,height=1.25in,clip,keepaspectratio]{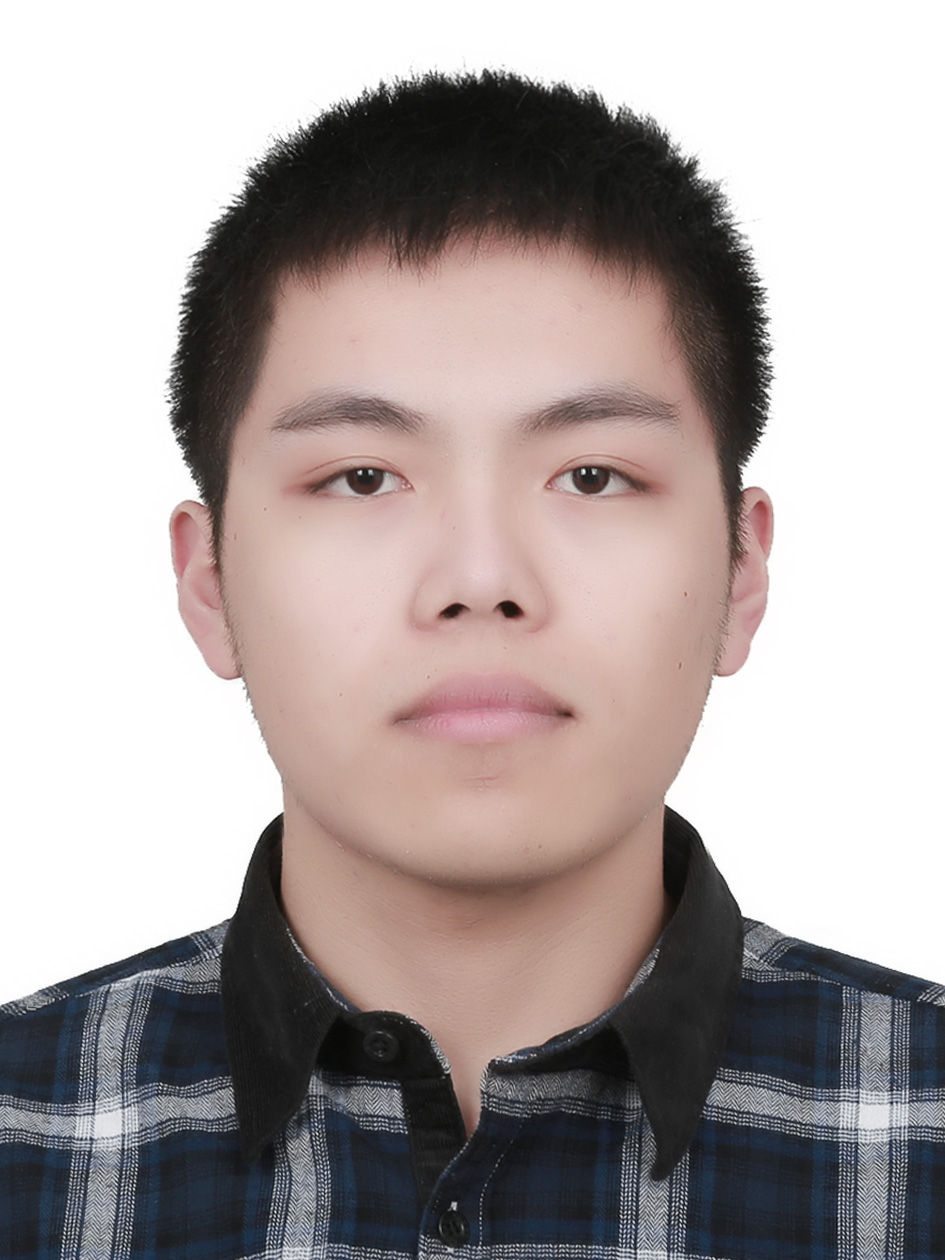}}]{Huan Mao} received his B.S. degree from the College of Physics and Information Engineering, Fuzhou University, in 2020. Currently, he is currently pursuing his master degree at the College of Physics and Information Engineering, Fuzhou University, Fuzhou, China. His research interests include computer vision and image processing.
\end{IEEEbiography}

\begin{IEEEbiography}[{\includegraphics[width=1in,height=1.25in,clip,keepaspectratio]{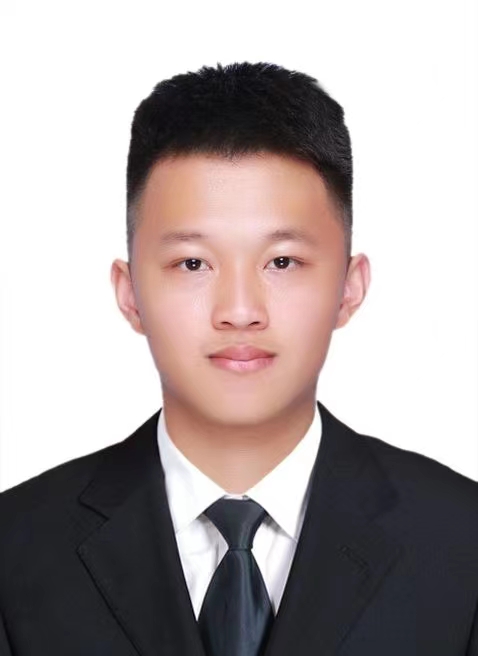}}]{Yulin Chen} received his B.S. degree from the College of Physics and Information Engineering, Fuzhou University, in 2021. Currently, he is currently pursuing his master degree at the College of Physics and Information Engineering, Fuzhou University, Fuzhou, China. His research interests include computer vision and image processing.
\end{IEEEbiography}

\begin{IEEEbiography}[{\includegraphics[width=1in,height=1.25in,clip,keepaspectratio]{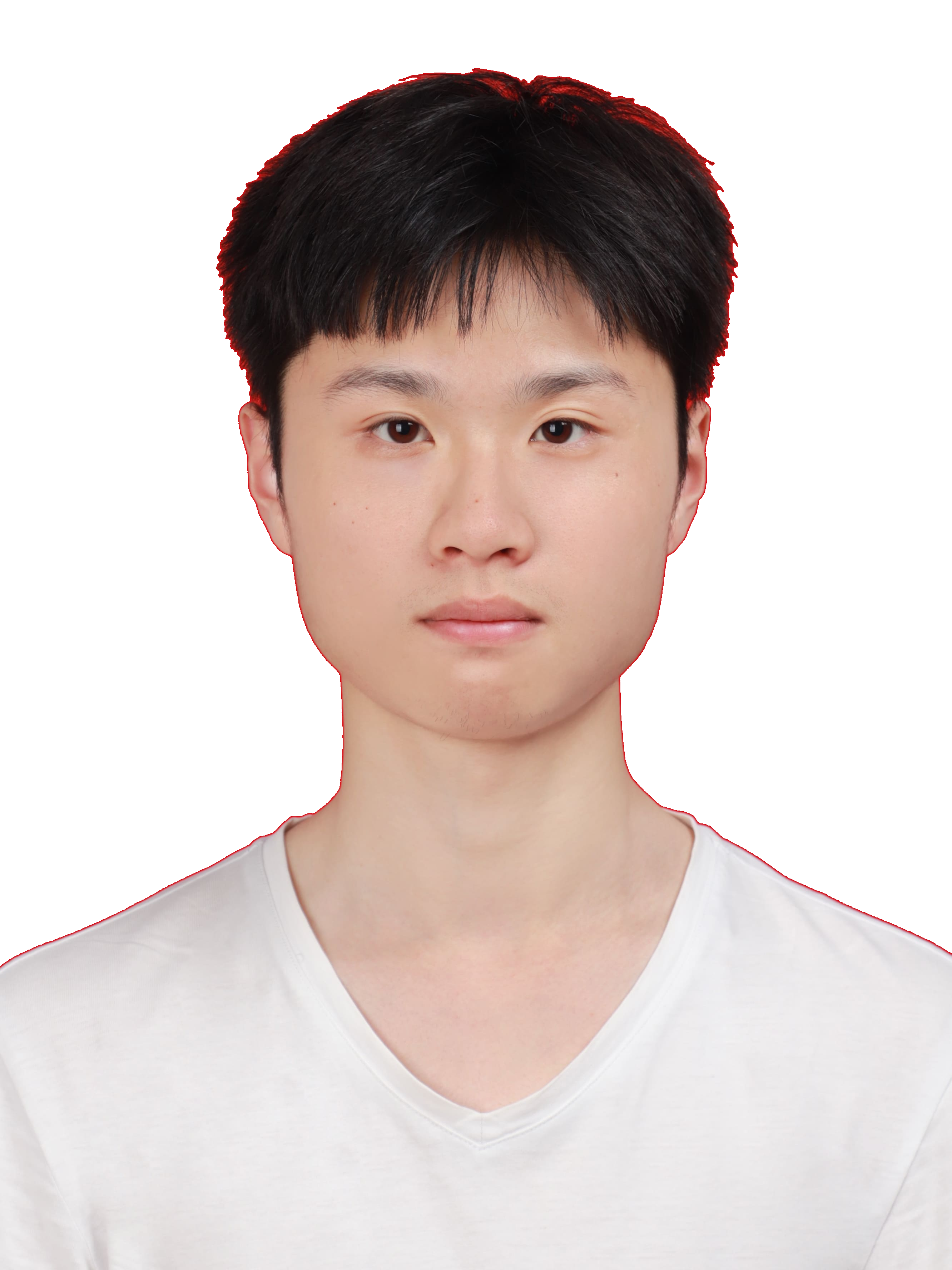}}]{Zongtan Li} currently pursuing his B.S degree at Maynooth Internetional Engineering College, Fuzhou University, Fuzhou, China. His research interests include computer vision, control theory.
\end{IEEEbiography}

\begin{IEEEbiography}[{\includegraphics[width=1in,height=1.25in,clip,keepaspectratio]{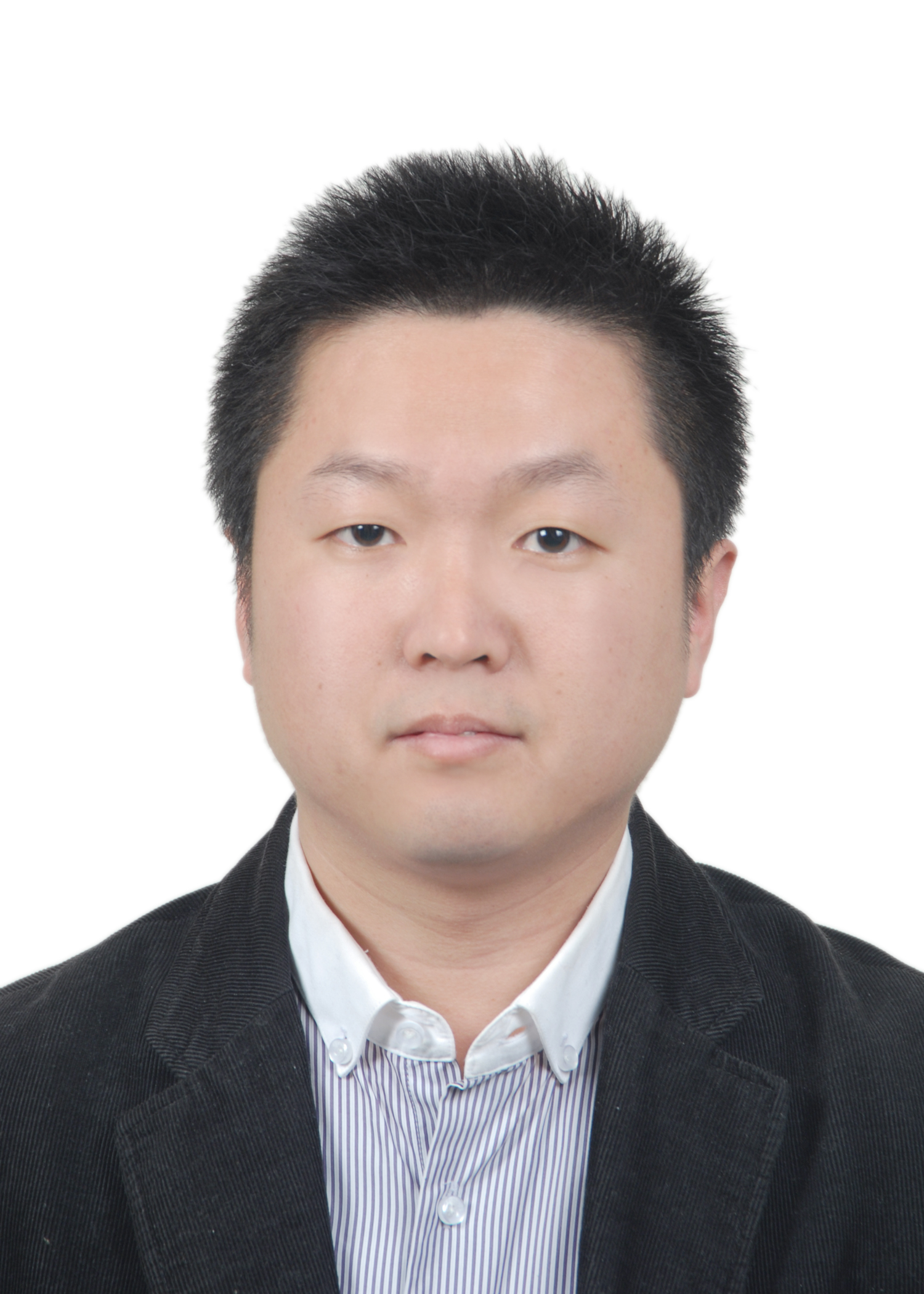}}]{Feng Chen} (Member, IEEE) was born in 1984. He received his Ph.D. degree in Signal and information processing at Beijing University of Posts and Telecommunications (BUPT), China, in 2014, and his M.S. degree in Communication Engineering from Guilin University of electronic technology (GUET), China, in June 2009. Currently, he is an assistant professor in College of Physics and information engineering in Fuzhou University, China. His current research interest includes radio resource management, heterogeneous networks and multimedia communication.
\end{IEEEbiography}

\begin{IEEEbiography}[{\includegraphics[width=1in,height=1.25in,clip,keepaspectratio]{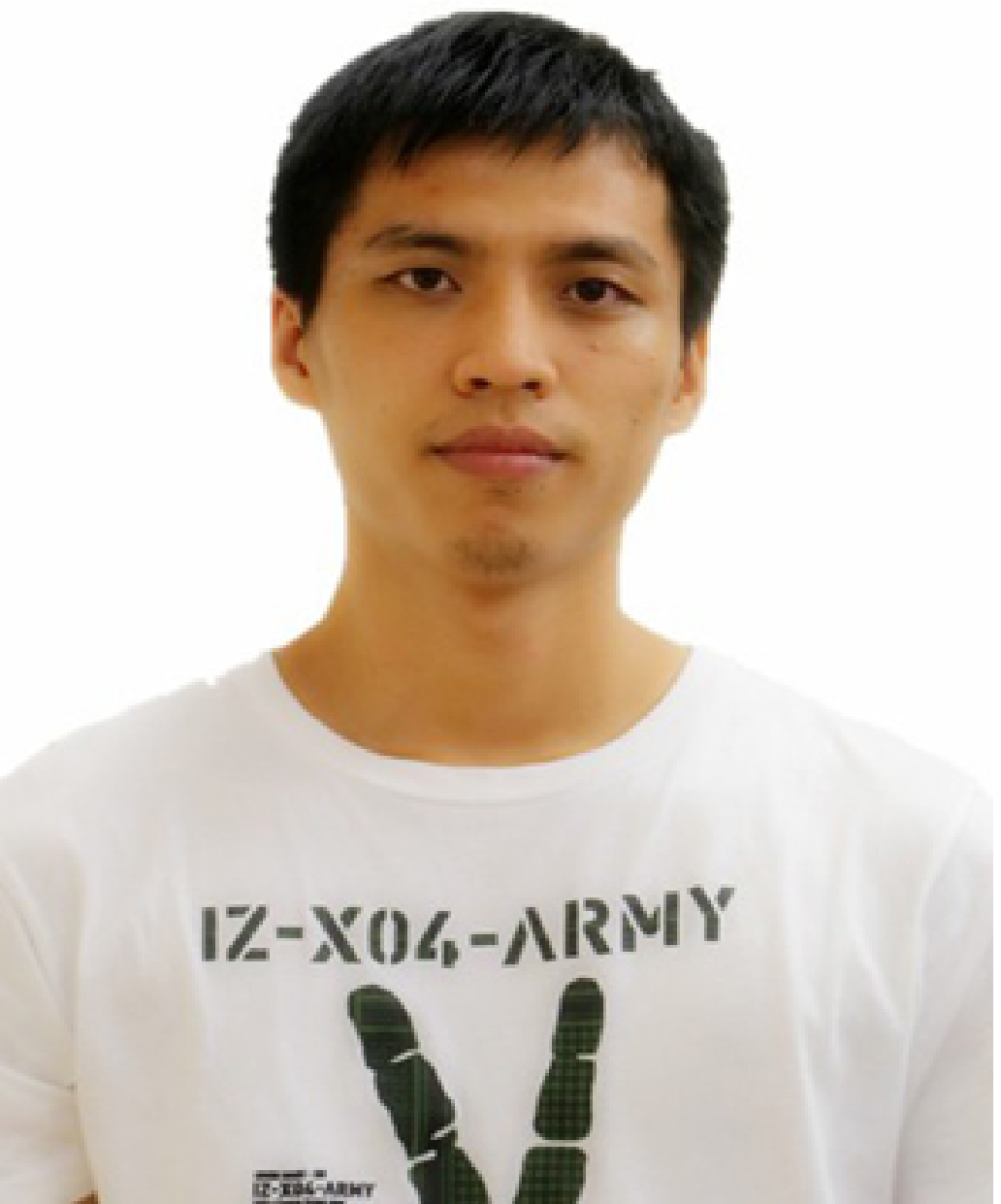}}]{Pingping Chen} (Member, IEEE) is currently a Professor in Fuzhou University, China. He received the Ph.D. degree in electronic engineering, Xiamen University, China, in 2013. From May 2012 to September 2012, he was a Research Assistant in electronic and information engineering with The Hong Kong Polytechnic University, Hong Kong. From January 2013 to January 2015, he was a Postdoctoral Fellow at the Institute of Network Coding, Chinese University of Hong Kong, Hong Kong. From July 2016 to July 2017, he was a Postdoctoral Fellow at Singapore University of Technology and Design. His primary research interests include machine learning,  image processing, and intelligient communications.
\end{IEEEbiography}

\end{document}